\documentclass[10pt,twocolumn,letterpaper]{article}

\usepackage{cvpr}
\usepackage{times}
\usepackage{epsfig}
\usepackage{graphicx}
\usepackage{amsmath}
\usepackage{amssymb}
\usepackage{mathtools}
\usepackage{booktabs}

\usepackage[pagebackref=true,breaklinks=true,letterpaper=true,colorlinks,bookmarks=false]{hyperref}

\def\cvprPaperID{2037} 

\DeclarePairedDelimiter\norm{\lVert}{\rVert}

\def\eg{\emph{e.g}\onedot}

\def\cenam{Center Attentive Module}
\def\scenam{Center-Attn}
\def\atm{Attention Transitive Module}
\def\satm{Attn-Trans}
\def\coram{Corner Attentive Module}
\def\scoram{Corner-Attn}
\def\aggatt{Aggregation Attentive Module}
\def\saggatt{Aggregation-Attn}

\cvprfinalcopy 

\def\cvprPaperID{****} 

\begin{document}

\title{SaccadeNet: A Fast and Accurate Object Detector}

\author{Shiyi Lan$^{1}$\thanks{This work was done when Shiyi Lan was a research intern at Wormpex AI Research.} ~~~Zhou Ren$^{2}$ 
~~~ Yi Wu$^{2}$
~~~ Larry S. Davis$^1$
~~~ Gang Hua$^{2}$\\
$^1$University of Maryland, College Park ~~~~~~~~~~ $^2$Wormpex AI Research\\
{\tt\small {sylan@cs.umd.edu,  lsd@umiacs.umd.edu}, {\{renzhou200622, ywu.china, ganghua\}}@gmail.com
}
}

\maketitle

\begin{abstract}
  
Object detection is an essential step towards holistic scene understanding. Most existing object detection algorithms attend to certain object areas once and then predict the object locations. However, neuroscientists have revealed that humans do not look at the scene in fixed steadiness. Instead, human eyes move around, locating informative parts to understand the object location. This active perceiving movement process is called \textit{saccade}. 

Inspired by such mechanism, we propose a fast and accurate object detector called \textit{SaccadeNet}. It contains four main modules, the \cenam, the \coram, the \atm, and the \aggatt, which allows it to attend to different informative object keypoints, and predict object locations from coarse to fine. The \coram~is used only during training to extract more informative corner features which brings free-lunch performance boost. On the MS COCO dataset, we achieve the performance of 40.4\% mAP at 28 FPS and 30.5\% mAP at 118 FPS. Among all the real-time object detectors, 
our SaccadeNet achieves the best detection performance, which demonstrates the effectiveness of the proposed detection mechanism.

\end{abstract}


\section{Introduction}

The human visual system is accurate and fast. As the first gate to perceive the physical world, our visual system glances at a scene and immediately understands what objects are there and where they are. This efficient and effective vision system enables human to perceive the visual world with little conscious thought. In machine intelligence, similarly a fast and accurate object detector is essential, which can allow machines to perceive the physical world efficiently and effectively, and unlock subsequent processes such as understanding the holistic scene and interacting within it.

\begin{figure}[t]
\centering
  \includegraphics[width=1.0\linewidth]{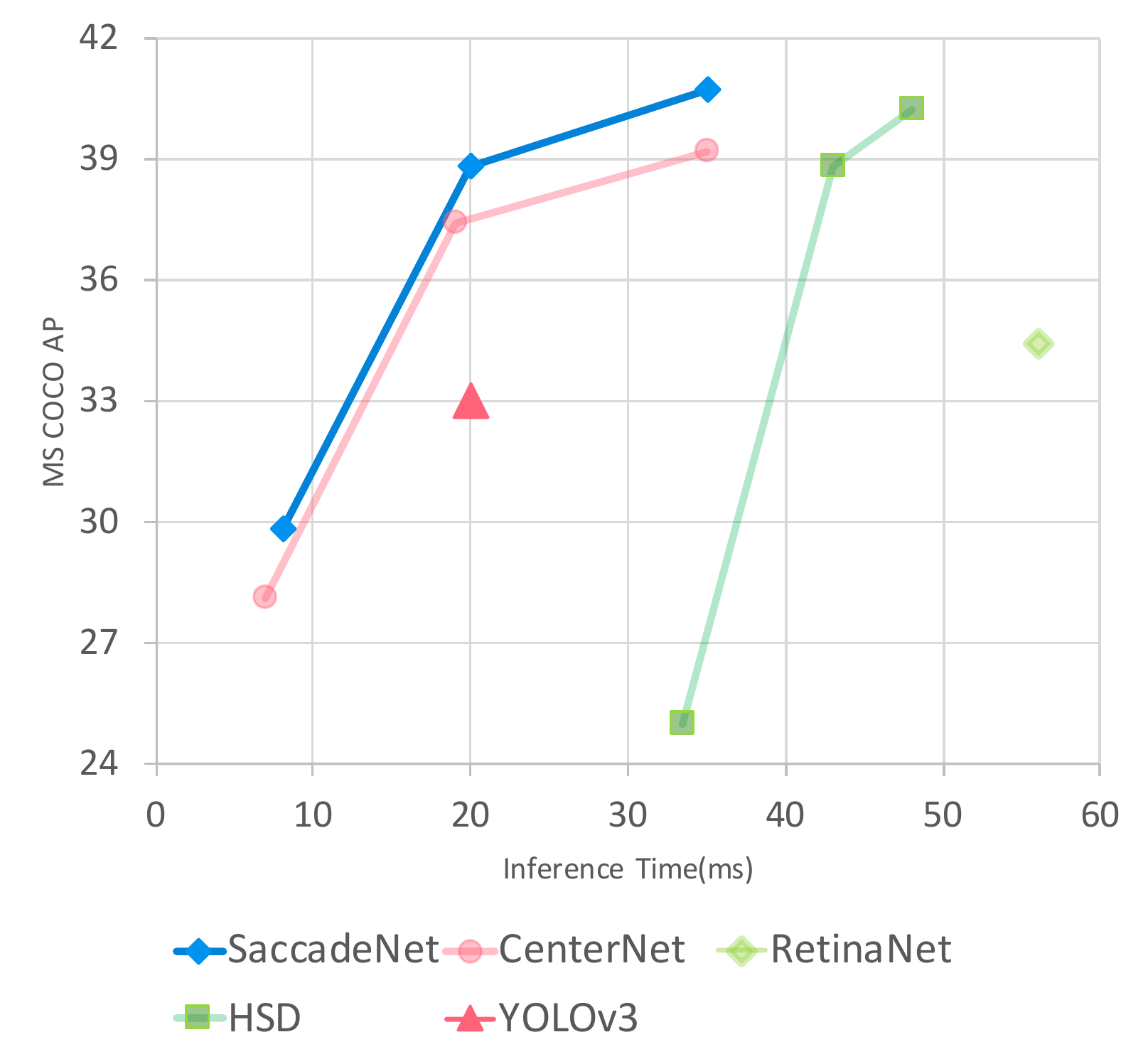}
  \\
  \caption{Performance comparison on COCO test-dev. SaccadeNet outperforms all previous fast detectors \cite{objectsaspoints, lin2017focal, cao2019hierarchical,yolo2016}. Best viewed in color.}
\label{fig:intro}
\end{figure}

Many recent algorithms have been proposed to advance object detection. On the one hand, anchor-based methods ~\cite{ren2015faster, redmon2018yolov3, lin2017focal,liu2016ssd,fu2017dssd} proposed to pre-define a large amount of anchor locations, and then either directly regress object bounding box locations, or generate region proposals based on anchors and decide whether each region contains a certain object category. These methods usually achieve competitive performance since they aggregate detailed image features within each region. However, the time-consuming region proposal stage is an bottleneck of inference speed.

On the other hand, researchers proposed anchor-free detectors ~\cite{law2018cornernet,zhou2019bottom,duan2019centernet, objectsaspoints}. This type of methods proposed to directly regress object locations by utilizing features at certain pre-defined object keypoints, either in the object center or on the bounding box edges. Most edge keypoints based methods are not fast because of the time-consuming grouping process that combines multiple detected keypoints to form a single object bounding box. The recent proposed center keypoint based detectors ~\cite{objectsaspoints} avoid the complex grouping process and run much faster. 

Most existing object detection algorithms steadily attend to certain object areas only once and then predict the object locations. During this one time of scanning for objects, different algorithms attend to different areas, either to the anchor boxes, to the proposed object regions, to the center keypoint, or to the edge keypoints. However, neuroscientists have revealed that~\cite{deubel1996saccade}, to understand an object's location, human do not look at the scene steadily. Instead, our eyes move around, locating informative parts to understand the object location. 

Inspired by such mechanism, we propose a fast and accurate object detector, named \textit{SaccadeNet}, which effectively attends to informative object keypoints, and predicts object locations from coarse to fine. Our SaccadeNet contains four main modules: the \cenam, the \coram, the \atm, and the \aggatt. The \cenam~predicts the object center location and category. Meanwhile, for each predicted object center, \atm~is used to predict the rough location of corresponding bounding box corners. To extract informative corner features, the \coram~is used to enforce the CNN backbone to pay more attention to object boundaries, so that the regressed bounding boxes are more accurate.  Finally, the \aggatt~utilizes the features aggregated from both the center and the corners to refine the object bounding boxes. 

SaccadeNet adopts multiple object keypoints including the center point and the corners, which encode and extract multiple levels of rich-detailed objects features. Moreover, it barely has speed loss comparing to the fastest center keypoint based detectors, since we predict object center and its corresponding corners jointly. Thus we do not need a grouping algorithm to combine them. Extensive experiments on the PASCAL VOC and MS COCO datasets have shown that SaccadeNet is fast and accurate. As shown in Figure \ref{fig:intro}, on COCO dataset when using ResNet-18 ~\cite{he2016deep,zhu2019deformable} as the backbone SaccadeNet achieves mAP of 30.5\% at 118 FPS. With DLA-34 ~\cite{yu2018deep}, SaccadeNet achieves 40.4\% mAP at 28 FPS, which is much better than other real-time detectors \cite{yolo2016,objectsaspoints}.

\section{Related Work}

Modern object detectors can be roughly divided into two categories: anchor-based object detectors and anchor-free object detectors. 

\subsection{Anchor-based Detectors}

After the seminal work of Faster R-CNN ~\cite{ren2015faster}, anchors have been widely used in modern detectors. It usually contains \textbf{two stages}.
The first-stage module is a region proposal network  (RPN), which estimates the objectness probabilities of all anchors and regresses the offsets between object boundaries and anchors. The second stage is R-CNN, which predicts the category probability and refines the boundary of bounding box. 

Recently, \textbf{anchor-based one-stage approaches}  ~\cite{redmon2018yolov3, lin2017focal,liu2016ssd,fu2017dssd} have drawn much attention in object detection because the architectures are simpler and usually run faster \cite{redmon2018yolov3}. They remove the RPN and directly predict the categories and regress the  boxes of candidate anchors. However, the performance of anchor-based one-stage detectors are usually lower than multi-stage detectors due to the extreme imbalance between positive and negative anchors during training. 

\subsection{Anchor-free Detectors}

Recently, anchor-free detectors have become more and more popular ~\cite{huang2015densebox,yolo2016, zhu2019feature,tian2019fcos,kong2019foveabox,liu2019high,objectsaspoints,law2018cornernet,duan2019centernet,zhou2019bottom,yang2019reppoints}. 
They avoid the complex design of anchors and usually run faster. 
The object detection is usually formulated as a keypoint detection problem so that the techniques of fully convolutional network (FCN) used in semantic segmentation \cite{Long2015fcn} and pose estimation \cite{Newell2016hourglass} can be applied for detection \cite{objectsaspoints}. 

YOLOv1~\cite{yolo2016} is one of the most popular anchor-free detectors. On each location of final layer of network, it predicts the bounding box, confidence of the box, and the class probability. In DenseBox \cite{huang2015densebox}, Huang \textit{et.al} extend the  FCN \cite{Long2015fcn} for face and car detection. The ground truth is a 5-channel map where the first one is a binary mask for the center of object and the other four are for the bounding box size.

After the seminal work of CornerNet~\cite{law2018cornernet}, 
\textbf{keypoint based anchor-free object detectors}
have drawn much attention. In CornerNet, the FCN directly predicts the corner heatmap, an embedding and a group of offsets for each corner. The embeddings are used to group the pairs of corner to form bounding boxes and the offsets remap the corners from low-resolutional heatmap to the high-resolutional input image. A corner pooling layer is proposed to better localize corners. ExtremeNet~\cite{zhou2019bottom} introduces a method that predicts the extreme points instead of the corners of bounding box, 
and the centerness heatmap is introduced for grouping step. 
In~\cite{duan2019centernet}, Duan \textit{et.al.} extend CornerNet by adding a center keypoint. The center keypoint is used to define a central region heuristically and then they use this region to refine the grouped corners.

\begin{figure*}[t]
\centering
  \includegraphics[width=0.9\linewidth]{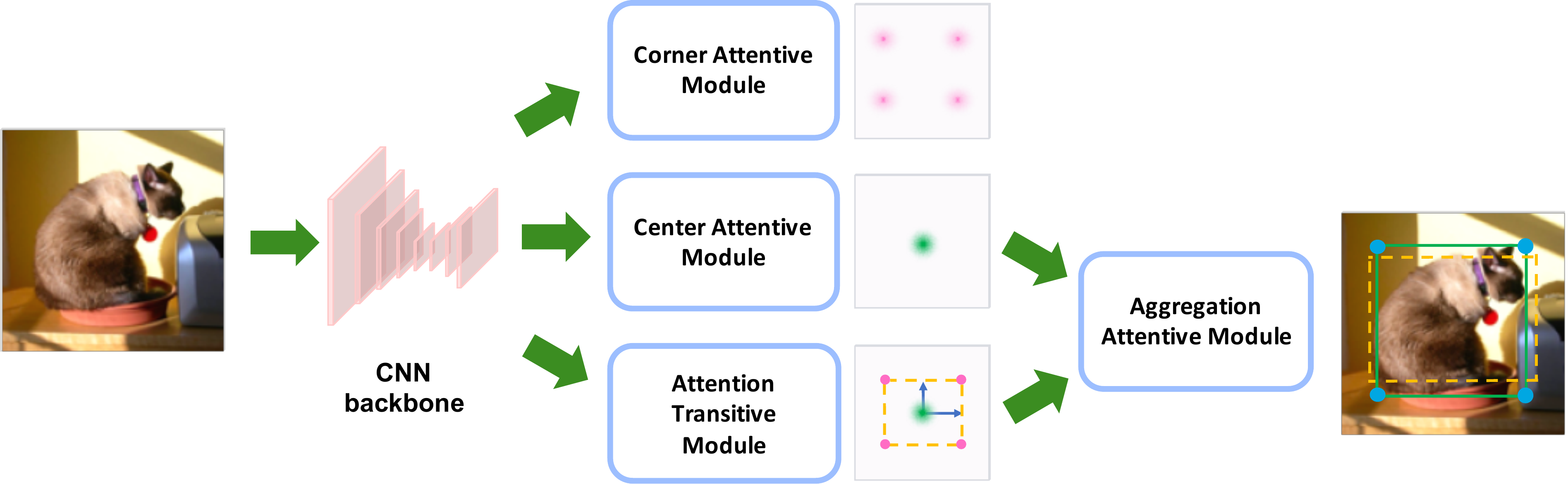}
  \\
  \caption{In SaccadeNet, we utilize 5 keypoints as informative parts for detection: the object center and 4 bounding box corners. After the CNN backbone, as in the middle branch, the \cenam~focuses at predicting the object center keypoint; then the \atm~in the bottom switches the attention from object center to estimate rough location of object corners. After that, the \aggatt~uses information aggregated from both center and corner keypoints, and predicts a refined location of objects. Moreover, in order to obtain informative corner features, the \coram~is used (in training only) to enforce the CNN backbone to pay more attention to object boundaries, as shown in the top branch.
  }
\label{fig:arch}
\end{figure*}

To avoid the complex grouping process, 
CenterNet
~\cite{objectsaspoints} directly predicts the center keypoint and the size of object.
Furthermore, it replaces IoU-based Non-Maximum Suppression (NMS) by peak keypoint extraction which can be run on GPU to reduce inference time. 
In ~\cite{tian2019fcos}  centerness is used to represent the objectiveness of the bounding box predicted at each location. In RepPoints ~\cite{chen2019revisiting}, a set of sample points is learned to bound the spatial extent of an object under the keypoint prediction framework.





\section{SaccadeNet}

It has been discovered that human eyes pick up informative parts to understand object locations instead of looking at every detail of objects~\cite{deubel1996saccade}, which makes it fast and accurate. To balance the trade-off between speed and accuracy, on top of the object center point, we use four object bounding box corner points as the informative keypoints in SaccadeNet since it naturally defines the bounding box position. SaccadeNet attends to these informative keypoints sequentially and then aggregates their features to infer accurate object locations. In this section, we will introduce four main modules of SaccadeNet respectively: the \cenam~(\scenam), the \atm~(\satm), the \aggatt~(\saggatt), and the \coram~(\scoram) used in training.




\subsection{Center Attentive Module}
\scenam~provides SaccadeNet the first sight of an object at its center and predicts object center keypoints. It takes the feature from CNN backbone as input and predicts the centerness heatmap. The centerness heatmap is used to estimate the categories and the center locations of all objects in the image. The number of channels in centerness heatmap is  the number of categories. 
Figure \ref{fig:arch} shows Center-Attn together with its output. 
In Center-Attn, it contains 2 convolutional layers. This 2-convolutional structure is called head module. It is a basic component for building other modules of SaccadeNet. We will describe it in details in Section~\ref{sec:exp}.

We use the Gaussian heatmap as ground truth  ~\cite{law2018cornernet}. 
The ground-truth heatmap for keypoints is not defined as either 0 or 1 because locations near the target keypoint should get less penalization than locations far away. 
Suppose the keypoint is at location $X_k$, the value at location $X$ on the ground-truth heatmap is defined as $e^{\frac{\norm{X - X_k}^2}{2 \sigma ^ 2}}$.
$\sigma$ is set to $1/3$ of the radius, which is determined by the size of objects to ensure that all locations inside the area could generate a bounding box with at least $t$ IoU with the ground-truth annotations. We follow the previous work ~\cite{law2018cornernet, duan2019centernet, objectsaspoints} and set $t$ as $0.3$.

Besides, a variant of focal loss ~\cite{lin2017focal} is applied to assist the Gaussian heatmap:
\begin{equation*}
L_{i,j}^{hm} = \\
\begin{cases} 
     (1-p_{i,j})^\alpha \log (p_{i,j}),  & \textit{if} \ y_{i,j} = 1\\
     (1-y_{i,j})^\beta  (p_{i,j})^\alpha log (1 - p_{i,j}) & \textit{otherwise}
\end{cases}
\end{equation*}
where $p_{i,j}$ is the score at location $(i,j)$ of heatmap and $y_{i,j}$ is the corresponding ground truth value.

\subsection{\atm}

\satm predicts the corners for all locations of the deep feature map. The output shape is $w_f \times h_f \times 2$ for a single image, where $w_f, h_f$ indicate the width and the height of feature map, respectively. The last dimension is designed to be 2 meaning the width and height of the bounding box. After we get the width and height of bounding box for each center at location $(i, j)$, we can compute the corresponding corners as $(i-w_{i,j}/2,j-h_{i,j}/2), (i-w_{i,j}/2,j+h_{i,j}/2), (i+w_{i,j}/2,j-h_{i,j}/2), (i+w_{i,j}/2,j+h_{i,j}/2)$.
In training, we adopt the L1 regression loss.
With \scenam~and \satm, SaccadeNet can generate object detections with coarse boundary. 

\subsection{\aggatt}

\saggatt~is proposed to attend to object center and bounding box corners again to predict a refined location. As shown in Figure~\ref{fig:arch}, it aggregates CNN features from corner and center keypoints using bilinear interpolation and outputs more accurate object bounding boxes. 
As shown in the experiments Section~\ref{sec:exp:iou-size}, \saggatt~is essential for us to obtain more accurate boundary.

\saggatt \  is a light-weight module for object boundary refinement.  Let $w_{i,j}, h_{i,j}$ indicate the width and height prediction at $ (i, j)$. Then, we calculate the corresponding top-left, top-right, bottom-left, bottom-right corners centering at position $ (i, j)$ by $ (i-w_{ i,j}/2, j-h_{ i,j}/2),  (i+w_{ i,j}/2, j-h_{ i,j}/2),  (i-w_{i,j}/2, j+h_{i,j}/2),  (i+w_{i,j}/2, j+h_{ i,j}/2)$. 
Since previous work~\cite{he2017mask} shows that bilinear sampling is helpful for the downsampled feature map, \saggatt \  takes the corners and center from the output of \satm, \scenam~and samples features from the backbone output by bilinear interpolation. 
The structure of \saggatt~ is a revised head module. We change the input of the first convolutional layer and let it take features of center and corners of object as input.

Finally, \saggatt~regresses the residual offsets to refine the boundary of objects by incorporating both the features from the corners and the center. The output of \saggatt~consists of residual width and residual height. We adopt L1 loss to train this module.

\subsection{\coram~in Training }



To extract informative corner features, we propose an auxiliary \scoram~branch (only in training) to enforce the CNN backbone to learn discriminative corner features. 
As shown in Figure \ref{fig:arch}, Corner-Attn uses one head module to process feature and output 4-channel heatmap including top-left, top-right, bottom-left, bottom-right corners.
Note that this branch is used only during training so that it is a free lunch for the increased inference accuracy.


The training of \scoram~is also based on the focal loss and Gaussian heatmap. We tried agnostic and non-agnostic heatmaps, meaning whether different object categories share the same corner heatmap output or not. In our experiments, there is no significant difference between their performance. For shorter training time and easier implementation, we use agnostic heatmaps for \scoram~in our experiments. 



\subsection{Relation to existing methods}


We will compare our work with other related work to address one of our contributions: SaccadeNet solves the issue of lacking holistic perception existed in edge-keypoint-based detectors and the issue of missing local details presented in center-keypoint-based detectors. 


\textbf{Edge-keypoint-based detectors} infer objects by assembling edge-keypoints, like corners \cite{law2018cornernet} or extreme keypoints \cite{zhou2019bottom}. They first predict edge keypoints and then use the grouping algorithm to generate object proposals. There are two possible problems that may make corner-keypoint-based fail to model holistic information: (a) Feature of corner encodes less holistic information since most corner-keypoint-based detectors~\cite{zhou2019bottom,duan2019centernet} still need feature of centers to assemble corner keypoints. (b) Corner keypoints often locate at background pixels which may encode less information than center keypoints do.  Although SaccadeNet also utilizes corner keypoints for bounding box estimation, it is still able to capture holistic by inferring bounding boxes directly from center keypoints. Meanwhile, SaccadeNet is very fast since it avoids the time-consuming grouping.


\textbf{Center-keypoint-based detectors} propose objects from center points \cite{objectsaspoints}. It outputs center heatmap and regresses boundary directly. However, center point may be far from the boundary of object so they may fail to estimate accurate boundary on some cases, especially for the large objects (as shown in Figure \ref{fig:results}). On the other hand, corner keypoints are naturally proximal to the boundaries, so it may encode more local accurate information. Lack of modeling corners may be harmful for the center-keypoint-based detectors. Therefore, SaccadeNet utilizes corner keypoints to alleviate this issue so that it can estimate more accurate boundary.  

SaccadeNet bridges the gap between edge-keypoint-based detectors and center-keypoint-based detectors.



\section{Experiments}\label{sec:exp}

The experiments are conducted on 2 datasets, PASCAL VOC 2012 ~\cite{everingham2010pascal} and MS COCO ~\cite{lin2014microsoft}. 
MS COCO dataset contains 80 categories, including 105k images for training (train2017) and 5k images for validation (val2017). 
Pascal VOC consists of 20 categories and it contains  a training set of 17k images and a validation set of 5k images. This setting is the same as previous work ~\cite{law2018cornernet, duan2019centernet, he2017mask, objectsaspoints}.

\subsection{Implementation}

\textbf{Backbone}. Our backbone consists of down-sampling layers and up-sampling layers. The down-sampling layers are from the CNN for image recognition, \eg ~\cite{yu2018deep,he2016deep}. The up-sampling layers use a couple of convolutional layers and skip connections to fuse high-level and low-level feature, \eg ~\cite{lin2017feature}. We choose DLA-34~\cite{yu2018deep} and ResNet-18 ~\cite{he2016deep} as the down-sampling backbone and use the up-sampling layers adopted in CenterNet~\cite{objectsaspoints}, where deformable convolutions \cite{zhu2019deformable} are used. The size of the backbone output is 1/4 of the input. The high-resolution output help SaccadeNet recognize and locate small objects. For fair comparison and to illustrate the effectiveness of SaccadeNet, we keep all the settings of backbone the same as \cite{objectsaspoints}.

\textbf{Head module}. The head module is the basic component of building four modules of SaccadeNet as illustrated in Figure \ref{fig:arch}. We use the unified structure of 2 convolutional layers for all the head modules. The first convolutional layer is followed by a ReLU layer with a kernel size of $3\times3$ and 256-dimension output channels. The second convolutional layer uses a $1\times 1$ kernel without activation function. 
Center-Attn contains one head module. The number of output channels of this module depends on the number of categories, \eg 20 for Pascal VOC, 80 for MS COCO. \scoram~contains one head module which outputs a 4-channel heatmap representing the agnostic heatmap of 4 corner keypoints.  
Corner-Attn contains 2 head modules with 2-channel output, indicating the two directional center offset and the width and height of object, respectively. \saggatt~contains one module with output of 2 channels denoting the residual offsets of width and height of object. The number of parameters of each head module is less than 200k.

\textbf{Training.} Our experiments were conducted on a machine with 4 GPUs of Geforce RTX 2080 Ti. It takes 10 days to train SaccadeNet-DLA34 and 5 days to train SaccadeNet-Res18. We use Adam ~\cite{kingma2014adam} for network optimization. For data augmentation, we apply random flipping, random scaling (range from 0.6 to 1.3), cropping and color jittering. On MS COCO dataset, the  size of input to the network is $512 \times 512$. We use a batch size of 32 (8 images on each GPU) with the initial learning rate of $1.25\times 10^{-4}$ for 210 epochs. The learning rate is dropped to $1.25\times 10^{-5}$ at the 181-th epoch.
The same training settings are used for CenterNet~\cite{objectsaspoints}.
We use different loss weights for the losses.  The loss weights for $L_{Corner-Attn}$, $L_{Center-Attn}$ and $L_{Aggregation-Attn}$ are 1, 1, 0.1, respectively. Corner-Attn outputs center offsets and the center-corner offsets. We use 0.1 for the loss weight of center-corner offsets and 1 for the loss weight of center offsets.
On PASCAL VOC 2012, we use a batch size of 32 on single GPU for training and the input shape of the network is $384 \times 384$. 
We set the initial learning rate to $1.25 \times 10^{-4}$ for 70 epochs. The learning rate is decreased to $1.25 \times 10^{-5}$, $1.25 \times 10^{-6}$ at the 46-th epoch, 61-th epoch, respectively. All the other settings are kept the same as our experiments on MS COCO dataset for training.
We use the parameters pretrained on ImageNet~\cite{deng2009imagenet} dataset to initialize the down-sampling layers. The parameters of up-sampling layers of backbone and head modules are randomly initialized.
\begin{table*}[]

\begin{tabular}{l@{\ }c@{\ \ \ }ccccccc}
\toprule
              & \textbf{Backbone}        & \textbf{FPS} & \textbf{AP} & $\textbf{AP}_{\textbf{50}}$ & $\textbf{AP}_{\textbf{75}}$  & $\textbf{AP}_{\textbf{S}}$   & $\textbf{AP}_{\textbf{M}}$   & $\textbf{AP}_{\textbf{L}}$   \\ \hline
TridentNet ~\cite{li2019trident} & ResNet-101-DCN  & 0.7 & \textbf{48.4} & 69.7 & 53.5 & 31.8 & 51.3 & 60.3 \\ 
SNIPER ~\cite{Singh2018SNIPER}   & DPN-98          & 2.5 & 46.1 & 67.0 & 51.6 & 29.6 & 48.9 & 58.1 \\
MaskRCNN ~\cite{he2017mask}      & ResNeXt-101     & 11  & 39.8 & 62.3 & 43.4 & 22.1 & 43.2 & 51.2 \\ \hline
RetinaNet ~\cite{lin2017focal}   & ResNeXt-101-FPN & 5.4 & \textbf{40.8} & 61.1 & 44.1 & 24.1 & 44.2 & 51.2 \\
YOLOv3 ~\cite{redmon2018yolov3}  & DarkNet-53      & 20  & 33.0 & 57.9 & 34.4 & 18.3 & 25.4 & 41.9 \\
HSD ~\cite{cao2019hierarchical} & ResNet101 & 21 & 40.2 & 58.2 & 44.0 & 20.0 & 44.4 & 54.9 \\
HSD ~\cite{cao2019hierarchical} & VGG16 & 23 & 38.8 & 58.2 & 42.5 & 21.8 &  41.9 & 50.2 \\
\hline
ExtremeNet ~\cite{zhou2019bottom}& Hourglass-104   & 3.1 & 40.2 & 55.5 & 43.2 & 20.4 & 43.2 & 53.1 \\
CornerNet ~\cite{law2018cornernet}& Hourglass-104   & 4.1 & \textbf{40.5} & 56.5 & 43.1 & 19.4 & 42.7 & 53.9 \\
CenterNet ~\cite{objectsaspoints}& DLA-34-DCN          & 52/28  & 37.4/39.2 & -/57.1 & -/42.8 & -/19.9 & -/43.0 & -/51.4 \\ 
$^*$CenterNet ~\cite{objectsaspoints}& ResNet-18-DCN     &  142/71 & 28.1/30.0 & 44.9/47.5 & 29.6/31.6 & -/- & -/- & -/- \\\hline
SaccadeNet                  & DLA-34-DCN         & 50/28  & 38.5/\textbf{40.4} & 55.6/57.6 & 41.4/43.5 & 19.2/20.4 & 42.1/43.8 & 50.6/52.8  \\
$^*$SaccadeNet               & ResNet-18-DCN    &   118/67  &  30.5/32.5 & 46.7/48.9      & 32.6/34.7  &  12.0/13.9  &   33.9/36.2  & 45.8/47.9  \\
\bottomrule \\
\end{tabular}
\caption{The experiments are conducted on MS COCO test-dev. SaccadeNet-DLA outperforms CenterNet-DLA by 1.2\% mAP with little overhead. This is the first detector that achieves more than 40\% mmAP on MS COCO test-dev with more than 25 FPS.  SaccadeNet-Res18 outperforms CenterNet-Res18 by 2.4\% mAP with small overhead. We show naive/flip testing results of CenterNet and SaccadeNet. A dash indicates the method doesn't provide the result. $^*$ means the experiments are conducted on MS COCO val2017. }
\label{tab:big_experiments}
\end{table*}


\begin{figure*}[t]
\centering
  \includegraphics[width=1.0\linewidth]{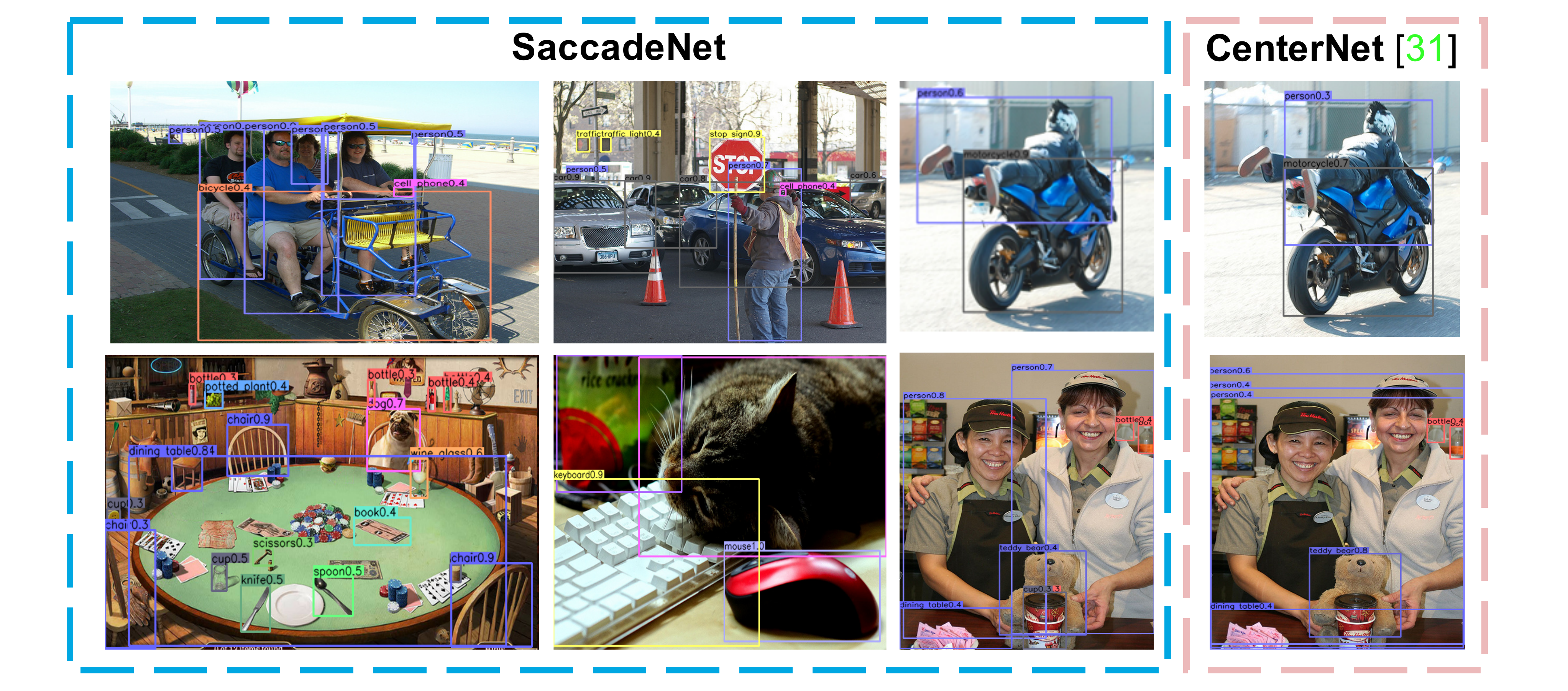}
  \\
  \caption{Qualitative Results of SaccadeNet and CenterNet \cite{objectsaspoints}. The images on the left 3 columns are the results of SaccadeNet-DLA34. The right column includes the results of CenterNet-DLA34\cite{objectsaspoints}. Best viewed in color. 
  }
\label{fig:results}
\end{figure*}

\textbf{Inference.} On MS COCO dataset, the size of input image is $512 \times 512$. Flipped testing is optional for better performance. When the flipped and the original images are both used as inputs, we average the outputs of Center-Attn, Corner-Attn, Aggregation-Attn. For higher speed, we use peak-picking NMS proposed in ~\cite{objectsaspoints} instead of IoU-based NMS for post-processing. Peak-picking NMS is a $3\times3$ pooling-like operator, which eliminates all non-peak activation. 
After NMS, we select the object proposals with top-100 centerness scores provided by Center-Attn. 
For Pascal VOC, we do not apply data augmentation for testing. We use Peak-picking NMS instead of IoU-based NMS. 

\begin{table*}[]
\centering
\begin{tabular}{ccccccc}
\toprule
                & \textbf{mAP@50}                   & \textbf{mAP@70}       & \textbf{mAP@90}       & \textbf{mAP@S} & \textbf{mAP@M}       & \textbf{mAP@L} \\ \hline
Baseline            &     70.69                     &   55.50               &  16.48                & 8.15                  & 23.74         & 57.86 \\ \hline
Corner-Attn     & \textbf{71.02/+0.33}              &  56.42/+0.92            &    13.51/-2.97      & \textbf{9.75/+1.60}   & 24.45/+0.71         & 58.84/+0.98   \\
Aggregation-Attn     &  70.64/-0.05                 &   55.85/+0.35           &   17.34/+0.86             &  8.30/+0.15     & 24.30/+0.56       & 58.39/+0.53 \\
Corner-Attn + Aggregation-Attn &    70.94/+0.25    &   \textbf{57.84/+2.34}     &  \textbf{21.07/+4.59}   & 9.69/+1.54      & \textbf{25.17/+1.43} & \textbf{60.40/+2.54}\\
\bottomrule 
\end{tabular}
\caption{This table shows the results of SaccadeNet with or without Aggregation-Attn and Corner-Attn. We use 6 metrics of different IoU thresholds and object sizes. 
All experiments are conducted on Pascal VOC. 
For our approaches, we show both the mAP and the mAP gain (+) or loss (-) compared with the baseline.}
\label{tab:iou-size}
\end{table*}

\begin{table}[]
\begin{tabular}{c@{\ \ \ }c@{\ \ \ }c@{\ \ \ }c@{\ \ \ }c@{\ \ \ }c}
\toprule
Backbone & Aggregation-Attn  & Flip & NMS & FPS & mAP \\ \hline
DLA      &      &      &  PP & 52  &   37.9   \\
DLA      & \checkmark  &    &  PP &  50   &  38.8    \\
DLA      &   & \checkmark     & PP & 28  &    39.9  \\
DLA      & \checkmark  & \checkmark  & PP   &  28   & 40.7     \\
DLA      & \checkmark  &   & IoU & 45 & 39.3 \\
DLA      & \checkmark & \checkmark & IoU & 27 & 40.9 \\
\bottomrule 
\end{tabular}
\caption{All experiments are conducted on MS COCO val2017. PP and IoU represent peak-picking NMS and IoU-based NMS, respectively.}
\label{tab:speed}
\end{table}

\subsection{Comparison with State-of-the-art Methods}

Table \ref{tab:big_experiments} shows the comparison results of our approaches with previous work. SaccadeNet achieves state-of-the-art performance with higher speed.

SaccadeNet-DLA34 achieves 40.4 mAP at 28 FPS. It outperforms CenterNet-DLA34~\cite{objectsaspoints} by 1.2\% AP without visible speed loss due to the light-weight head modules. Besides, our approach outperforms the classic two-stage detector, MaskRCNN~\cite{he2017mask}. Meanwhile, we achieve approximately 3 times speed of it. 
Compared with RetinaNet~\cite{lin2017focal}, SaccadeNet-DLA34 performs approximately 4 times faster 
with only 0.4\% drop in accuracy. 
As shown in Table \ref{tab:big_experiments}, SaccadeNet-DLA34 is faster and much more accurate than YOLOv3~\cite{redmon2018yolov3}. 
We compare the results of SaccadeNet-DLA34 and CenterNet-DLA34~\cite{objectsaspoints} with different IoU thresholds and of different sizes. The average precision gains $+0.5$, $+0.7$ of IoU@0.5, IoU@0.7 and gains $+0.5$, $+0.8$, $+1.4$ of objects with small, medium, large size, respectively. SaccadeNet benefits more for high-IoU and large object proposals than others.  We will study how \saggatt~and \scoram~affect the object proposals of different quality and various size in Section \ref{sec:exp:iou-size}. Figure \ref{fig:results} shows the qualitative results of SaccadeNet and CenterNet. 
With the help of Aggregation-Attn, SaccadeNet is able to locate more accurate boundaries of objects.

Another version of our approach is based on ResNet-18 with deformable convolutions. SaccadeNet-Res18 is the first real-time anchor-free detector that achieves more than 30\% mAP on MS COCO val2017 with speed faster than 100 FPS.


\subsubsection{Efficiency Study}
We will discuss 4 main factors of efficiency: backbone, head modules, data augmentation, non-maximum suppression. 

\textbf{Backbone}. We use DLA-34 ~\cite{yu2018deep} and ResNet-18 ~\cite{he2016deep} with additional up-sampling layers used in CenterNet~\cite{objectsaspoints} as backbone. DLA-34 runs at 18.4 ms per image. ResNet-18 runs at 6.8 ms per image. The total inference time of SaccadeNet with DLA-34 and ResNet-18 is 20 ms, 8.5 ms per image, respectively. The efficiency of backbone is the major bottleneck of speed.

\textbf{Head modules}. There are $64\times 256\times 3 \times 3 + 256\times C_{out}$ parameters for each head module, where $C_{out}$ denotes the number of output channels. There are only 3 head modules during inference. The largest head module is the predictor of Center-Attn, which only contains 168k parameters. The only concern is that the inputs of \saggatt~depend on the outputs of Center-Attn and Corner-Attn. It may cause sequential execution that may increase the inference time. Fortunately, the execution turns out to be very fast. The inference time of all the head modules is much smaller than the backbone, which only cost 1.5 ms and 1.6 ms for SaccadeNet-DLA34 and SaccadeNet-Res18. The performance of SaccadeNet with and without \saggatt~is illustrated in Table \ref{tab:speed}. Obviously, \saggatt~is important for the performance improvement.

\textbf{Data augmentation}. For better performance, we feed the network with both the flipped image and the original image. Although this technique will double the inference time theoretically, it significantly improves the performance. Figure \ref{tab:speed} illustrates the performance of SaccadeNet with and without flip testing.

\textbf{Non-maximum Suppression}.  In SaccadeNet, we replace the popular IoU-based NMS with peak-picking NMS. Peak-picking NMS performs $3\times 3$ pooling on the output heatmap of Center-Attn. The inference time of it is less than 0.1ms. In comparison, the IoU-based NMS needs 2 ms for post-procession.  Table \ref{tab:speed} shows the comparison between IoU-based NMS and peak-picking NMS.

\subsection{Ablation Study}

In this section, we will study the characteristics of SaccadeNet. We conduct the experiments with SaccadeNet-Res18 on Pascal VOC.

\textbf{Evaluation metrics}. For detailed evaluation, we use 6 metrics for different IoU thresholds and size: AP@50, AP@70, AP@90, AP@S, AP@M, AP@L. AP@50, AP@70, AP@90 represent the average precision using IoU thresholds of 50\%, 70\%, 90\%, respectively. For evaluating objects of different size, we define AP@S, AP@M, AP@L as the average precision of small objects, medium objects, and large objects. Small, medium, large objects contain objects with area of $[0,64^2]$, $[64^2, 128^2]$, and $[128^2, \infty]$, respectively.

\subsubsection{Benefits of Aggregation-Attn and \scoram}\label{sec:exp:iou-size}
Our proposed Aggregation-Attn and \scoram~are designed to improve the quality of boundary. 
To study how much they affect high-quality/low-quality and large/small object proposals, we use different IoU thresholds to compute the mean average precision and evaluate it on the objects of different sizes. 
As shown in Table \ref{tab:iou-size}, larger objects and high-quality bounding boxes gain more benefits with Aggregation-Attn and Corner-Attn. 


\subsubsection{Keypoint Selection}

Although our proposed SaccadeNet reveals that corners are very important for accurate boundary localization, it is still unknown whether other keypoints are helpful for bounding box regression. We try different kinds of points: middle-edge points and other inner-box points.

\begin{figure}[t]
\centering
  \includegraphics[width=0.8\linewidth]{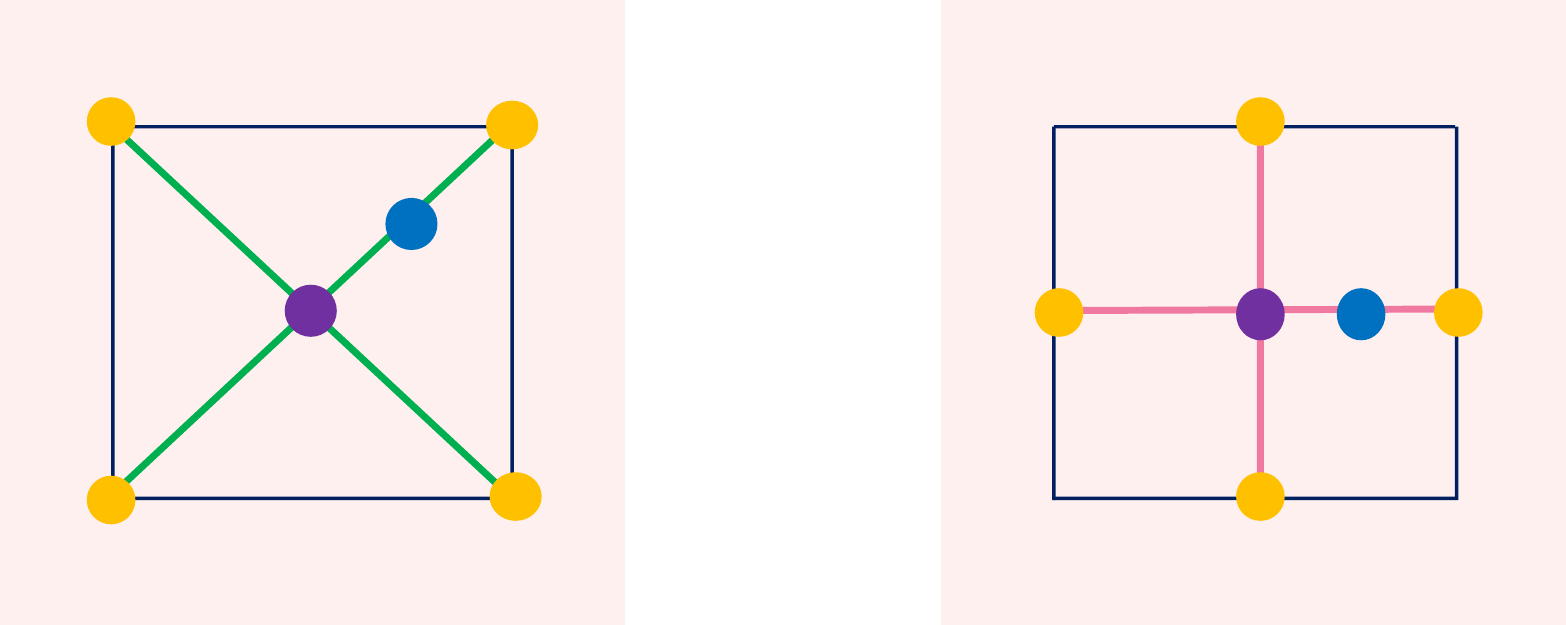}
  \\
  \caption{Purple points and yellow points denote centers and corners, respectively. On the left, the green lines denote the diagonal lines of bounding box. The blue point represents a Diag Pts computed by bilinear interpolation. On the right, the yellow points are middle points of bounding-box sides. The pink lines denote the middle line of the bounding box. The two end of middle line are two opposite yellow points. The blue point represents a Mid-edge Pts.}
\label{fig:corvsotherpoints}
\end{figure}

The middle-edge points of an object are the 4 points in the middle of 4 edges of a bounding box. We also replace corners with points on the orthogonal lines of the bounding box. Figure \ref{fig:corvsotherpoints} describes the keypoints mentioned above. We change the corners to other keypoints as inputs of Aggregation-Attn and the annotations from corners to other keypoints for Corner-Attn. Table \ref{tab:corvsotherpoints} illustrates the results on Pascal VOC. 

We find that the corners are the most helpful keypoints for SaccadeNet among all other keypoints except centers. We also find that keypoints closer to corners leads to higher performance for both Aggregation-Attn and Corner-Attn.  
One possible reason is that corners define the extent of the object and we use the bounding box for loss calculation.

\begin{table}[]
\begin{tabular}{c@{\ \ }ccc}
\toprule
                                         & \textbf{mAP@50}               & \textbf{mAP@70}               & \textbf{mAP@90}               \\ \hline
Corners                                  &        \textbf{70.94}              &          \textbf{57.84}            &     \textbf{21.07}                 \\ \hline
Diag Pts@0.8                          &          70.92            &      57.32                &        18.27              \\
Diag Pts@0.6                          &             70.59         &           56.48           &    17.40                  \\
Diag Pts@0.4                          &            70.43          &          56.11            &     17.31                 \\ 
Mid-edge Pts@1.0               &        70.64                  &          55.85              &     17.34                 \\
Mid-edge Pts@0.8 &                70.43    &        55.33              &  17.29 \\
Mid-edge Pts@0.6 &               70.51     &         55.10             &  16.98 \\
\bottomrule
\end{tabular}
\\

\caption{This table shows the results of using different points for Corner-Attn on PASCAL VOC with ResNet-18. Corner represents the original SaccadeNet-Res18. Diag Pts@t (t is a float number) represents the points locating at $p_{ct} * (1-t) + p_{cr} * t$, where $p_{ct}$, $p_{cr}$ represents the position of centers and corners. Similarly, Mid-edge Pts@t represents a points locating at $p_{ct} * (1-t) + p_{ml} * t$, where $p_{ct}$ and $p_{ml}$ indicate center points and middle points of an edge of object bounding box. Figure \ref{fig:corvsotherpoints} describes the position of all points mentioned above.}
\label{tab:corvsotherpoints}
\end{table}

\subsubsection{Does Iterative Refinement Help?}

An intuitive idea for improving SaccadeNet is to apply Aggregation-Attn iteratively. In the experiments, we use a couple of sequential modules of Aggregation-Attn. The outputs of the previous module are used as inputs in the next module. Table \ref{tab:cas_ref} shows the results on PASCAL VOC. 

The results show that iterative refinement works for more accurate boundary. The finer bounding boxes get more improvement by iterative refinement. However, as a result of more sequential execution, the iterative refinement is not very efficient. Due to speed-accuracy trade-off, we only use one Aggregation-Attn in all the other experiments.

\begin{table}[]
\begin{tabular}{c@{\ \ \ \ \ \ \ \ \ \ \ }ccc}
\toprule
Num of iter. & \textbf{mAP@50} & \textbf{mAP@70} & \textbf{mAP@90} \\ \hline
0            &  71.02  &  56.42 &   18.96  \\
1            &  70.94  &  57.84 &   21.07 \\
2            &  71.09  &  58.18 &   \textbf{21.32} \\
3            &  \textbf{71.12}  &  \textbf{58.42} &   20.70 \\
\bottomrule
\end{tabular}
\caption{The table shows the results of applying iterative refinement on SaccadeNet with different IoU thresholds. All the experiments are based on ResNet-18 on PASCAL VOC. Num of iter means the number of iterations used for boundary refinement}
\label{tab:cas_ref}
\end{table}

\subsubsection{Does Aggregation-Attn also Help Classification?}

Object detection is the step to understand ``what is where". We have validated that Aggregation-Attn improves the localization of object by fusing feature of corner and center keypoints, namely it helps in terms of ``where". Now we want to study whether such information aggregation also helps in terms of ``what". We add another module, namely Aggregation Attentive Classifier (Aggregation-Attn-Cls) to refine classification scores. Its structure is the same as \coram. We use the classification scores to replace the original object classifier output.  Table \ref{tab:cls_ref} illustrates the results. Unfortunately, the performance is degraded by Aggregation-Attn-Cls. One possible reason is that the feature of corner keypoints encode little high-level discriminative information for classification.

\begin{table}[]
\centering
\begin{tabular}{c@{\ \ }c@{\ \ }c@{\ \ }c}
\toprule
Aggregation-Attn-Cls & \textbf{mAP@50} & \textbf{mAP@70} & \textbf{mAP@90} \\ \hline
                          &    \textbf{70.92}    &     \textbf{57.49}   &  18.96      \\
\checkmark                          &     52.26   &    43.23    &   \textbf{19.80}    \\
\bottomrule
\end{tabular}
\caption{This table shows the results of using Aggregation-Attn-Cls for classification with different IoU thresholds. All Experiments are performed on Pascal VOC with ResNet-18.}
\label{tab:cls_ref}
\end{table}



\subsubsection{Impact of the Center and Corner Keypoints in Aggregation-Attn module}

The experimental results in Section \ref{sec:exp:iou-size} have shown that the aggregation of features from corners and center in Aggregation-Attn is of great importance for the performance improvement.
However, is the feature fusion of the corners and center necessary and helpful? How much improvement does it gain by using center-only or corner-only feature?

To address these questions, we change the inputs of Aggregation-Attn into feature of center keypoints or feature of corner keypoints. Table \ref{tab:contri} shows that it is useful to fuse feature of corner and center keypoints together. Comparing to the first row where Aggregation-Attn module is not used, by using the center feature alone it barely improves the performance since previous Center-Attn module already use center feature. By using corner features alone, the performance is improved significantly. By incorporating feature of both corner and center keypoints, the detection result is further improved, especially in high-IOU thresholds.

\begin{table}[]
\begin{tabular}{ccccc}
\toprule
Corner     & Center     & \textbf{mAP@50}         & \textbf{mAP@70} & \textbf{mAP@90} \\ \hline
           &            & 71.02          & 56.42  & 18.96  \\
           & \checkmark & 70.89          & 56.55  & 19.01  \\
\checkmark &            & \textbf{71.04} & 57.53  & 19.78  \\
\checkmark & \checkmark & 70.94          & \textbf{57.84}  & \textbf{21.07} \\
\bottomrule
\end{tabular}
\\
\caption{This table shows the results of using different inputs for Aggregation-Attn with different IoU thresholds. All Experiments are performed on Pascal VOC with ResNet-18.}
\label{tab:contri}
\end{table}

\section{Conclusion}
We introduce {SaccadeNet}, a fast and accurate object detection algorithm. Our model actively attends to informative object keypoints from the center to the corners, and predicts the object bounding boxes from coarse to fine. SaccadeNet runs extremely fast, because these object keypoints are predicted jointly so that we do not need a grouping algorithm to combine them. We extensively evaluate {SaccadeNet} on PASCAL VOC and MS COCO datasets, which both demonstrates its effectiveness and efficiency.

\section{Acknowledgement}

Gang Hua was supported partly by National Key R\&D Program of China Grant 2018AAA0101400 and NSFC Grant 61629301. We deeply appreciate the help of Xingyi Zhou and Zuxuan Wu.

\vspace{3cm}

{\small
\bibliographystyle{ieee_fullname}
\bibliography{egbib}
}

\end{document}